# Segmentation of Bleeding Regions in Wireless Capsule Endoscopy Images: an Approach for inside Capsule Video Summarization

Mohsen Hajabdollahi, Reza Esfandiarpoor, S.M.Reza Soroushmehr,
Nader Karimi, Shadrokh Samavi, Kayvan Najarian

*Abstract*— **Wireless capsule endoscopy (WCE) is an effective means of diagnosis of gastrointestinal disorders. Detection of informative scenes by WCE could reduce the length of transmitted videos and can help with the diagnosis. In this paper we propose a simple and efficient method for segmentation of the bleeding regions in WCE captured images. Suitable color channels are selected and classified by a multi-layer perceptron (MLP) structure. The MLP structure is quantized such that the implementation does not require multiplications. The proposed method is tested by simulation on WCE bleeding image dataset. The proposed structure is designed considering hardware resource constrains that exist in WCE systems.**

***Keywords: Wireless capsule endoscopy; neural network quantization; hardware implementation.***

## I. INTRODUCTION

Wireless capsule endoscopy (WCE) is a non-invasive, painless endoscopic method which can be used in screening all parts of gastrointestinal (GI) for diagnostic and other medical experiments [1]. WCE imaging is always used for patients suspicious to bleeding and other types of abnormalities such as Crohn's disease in GI. Also it is possible to use WCE for patients with polyposis syndrome and small bowel disorder [1]. Since practical usage of WCE in 2001, researchers investigated the automatic methods for detection of abnormalities. These methods reduce the time spent by physician for medical diagnostic and abnormality detection [2].

In recent years, researchers investigated the problem of detection of ulcer, bleeding, polyps and other abnormalities in WCE images. In [3], a bleeding detection method is proposed consisting of feature extraction and classification steps. Histograms of K-Means clustering are used as features and these features are classified by convolutional neural network (CNN). Fu *et al.* classify super pixel color features using support vector machine (SVM) to speed up the bleeding detection process [4]. In [5], to avoid miss-classification of bleeding regions, noise and edges are removed and then dark and light regions are removed. A combination of RGB channels at block level is used for SVM classification. Jia and Meng employed a CNN structure to detect bleeding frames [6] where dataset for training is balanced in terms of bleeding images and non-bleeding ones. To produce more images for training CNN, datasets are augmented by rotation and mirroring. In [7], bleeding images are classified into bleeding and non-bleeding by an SVM. After that bleeding regions are segmented by CNN. In [8], local binary pattern (LBP) feature is extracted from channel I of HIS color space in 8×8 image blocks. Then each block is segmented by SVM. In [9], a method based on cellular automata is presented for clustering the WCE images. Classification of the cluster centroids by SVM is utilized for segmentation of bleeding regions. In [10], histogram of the pixels is assigned to each cluster in Ycbcr color space, then it is used as descriptor and by using an SVM, WCE images are segmented. In [11], super pixel saliency in different scales is obtained. Saliency is calculated based on color and texture information. For texture representation LOG-Gabor filter, SIFT and LBP are used. For color information, second channels of HIS and CMYK color spaces are used. All features are coded with K-means clustering and final saliency map is obtained by max-pooling on the saliency map and coded features.

In [12], a saliency based video summarization technique using different-order of moments is proposed. In [13], a new method for detection of polyp and ulcer in WCE frames is presented. Gabor filter for simulating the human visual system is applied and for improving the accuracy of the detection algorithm, edges are detected using SUSAN edge detection method. Then using the results from previous stages, ulcer and polyp regions are detected base on SVM and a decision tree. In [14], different textures including LBP, SIFT and HOG are used around the image key points. The set of extracted features is clustered and classified using SVM. In [15], a new method for detection of abnormalities in WCE images is presented. Features of image patches, including color and texture, are extracted using moments, LBP and DWT and they are made into two histograms. Importance of the features is identified using saliency map and based on this importance, WCE images are classified by SVM.

In WCE image analysis, only a few of the images are informative. Hence, there is no need to transmit every frame that is captured by the capsule. Image analysis and informative frame detection, due to limited memory capacity and power constraints, are very important [12]. There is a great demand for implementation of diagnostic process inside WCE. In this regard, Khorsandi *et al.* designed a hardware

M. Hajabdollahi, R. Esfandiarpoor, and N. Karimi are with the Department of Electrical and Computer Engineering, Isfahan University of Technology, Isfahan 84156-83111, Iran.
S.M. R. Soroushmehr is with the Department of Computational Medicine and Bioinformatics and the Michigan Center for Integrative Research in Critical Care, University of Michigan, Ann Arbor, MI, U.S.A.
S. Samavi is with the Department of Electrical and Computer Engineering, Isfahan University of Technology, Isfahan 84156-83111, Iran. He is also with the Department of Emergency Medicine, University of Michigan, Ann Arbor, MI, U.S.A.
K. Najarian is with the Department of Computational Medicine and Bioinformatics; Department of Emergency Medicine; and the Michigan Center for Integrative Research in Critical Care, University of Michigan, Ann Arbor, MI, U.S.A.

architecture for WCE image assessment inside the capsule [16]. In their architecture, informative frames are detected inside the capsule using mean, variance, skewness and kurtosis. Image compression inside WCE is introduced as another way to lower the power consumption. In [17], hardware architecture for image compression in WCE image is proposed. To compress the images, an integer based DCT transform as well as efficient coefficient encoding with low complexity is applied. In [18], a hardware core including camera interface, FIFO queue, controller and image compressor is presented. In [19], a new architecture to calculate DCT transform in image compression inside WCE is designed. The DCT transform that is used, is based on 1-D DCT transform which reduces the number of add and shift operations.

In this paper we present a new method for bleeding detection and segmentation in WCE. The proposed method has low complexity and it can be implemented inside the capsule. Color features are analyzed to select channels with more information about bleeding regions. For simplification purposes, an artificial neural network (ANN) is used which works on the three chosen color spaces. More simplification of ANN is achieved by using ternary quantization of the connections during the training.

The reminder of this paper is organized as follows. In Section II, proposed method for segmentation of bleeding regions in WCE images is demonstrated. Section III is dedicated to experimental results. Finally conclusion remarks are presented in section IV.

## II. PROPOSED METHOD

In Fig. 1, overview of the system for bleeding detection is illustrated. From a frame of WCE video, three color spaces, which have the most information, are extracted. Due to the limited power capability and other hardware constraints, a multi-layer perceptron (MLP) structure is applied. For more simplification and embedding the overall system inside capsule endoscopy, the network connections are quantized. In the following step the proposed system is explained in more details.

### A. Color space conversion

Colors can be represented in different spaces and in each color space, some features are more representative. For example in retinal image analysis, the green channel is more useful than the others. Color space conversion can be considered as a preprocessing step before any medical image processing. HSV and CIE lab color spaces are two alternatives of RGB color space. CIE lab is more intuitive than RGB and is designed to approximate the human vision system. L, "a" and ""b components represent color and L lies in the range of 0-100 and both "a" and "b" lie in -110-110. HSV color space including hue, saturation and value is quite similar to the way in which the human perceives color. In Fig. 2, a bleeding frame of WCE video is represented in different color spaces. As illustrated in Fig. 2, some color spaces represent the bleeding regions better than the others and some of them have no useful information. By a simple experiment, it is possible to identify which color spaces or channels are suitable to represent the bleeding regions. Each pixel value in each channel is considered as an index of a look up table (LUT). The content of LUT in the corresponding indices is increased for bleeding regions in all images of the dataset. By testing the contents of the LUT in bleeding regions it is possible to find the most suitable color space or channel. By this experiment, saturation, "a" component and gray scale representation, are selected as the best color features. Also it is observed from Fig. 2, that gray-scale conversion, saturation, and the "a" channel have distinctive information about the bleeding region.

### B. Neural Network for local classification of the bleeding regions

As illustrated in Fig. 2, an MLP structure is applied for classification of the WCE image regions. Around each pixel a patch is considered. Extracted patches from three color spaces are aligned as the input of the MLP. For each patch in the training phase, class of the central pixel is considered as the output of the network. Similar to other binary classification problems, two neurons are considered as the output of the network and a sotfmax operation makes the final class of the input patch. Considering resource limitations of the WCE, no further feature extraction and preprocessing has been applied. Experiments on WCE images show that only 0.2% of the pixels belong to the bleeding class and the rest are non-bleeding which is known as imbalance data. This problem is solved in the training

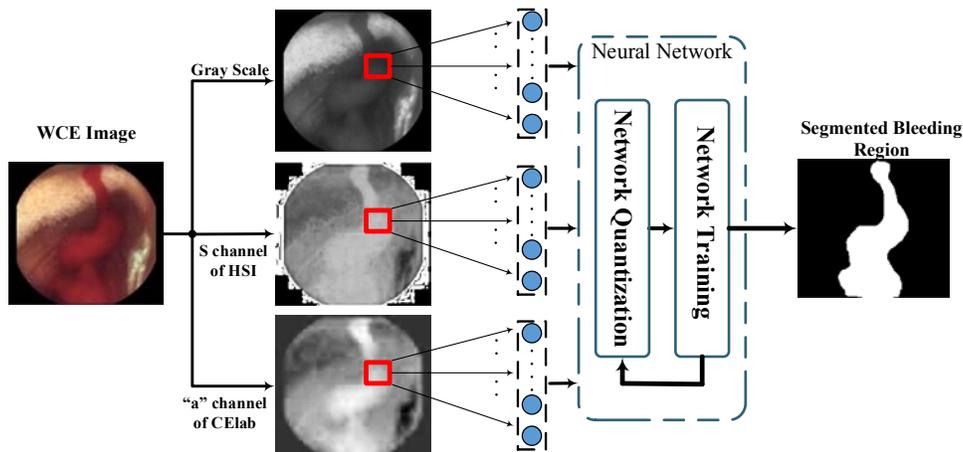

Figure 1. Overview of the proposed system.

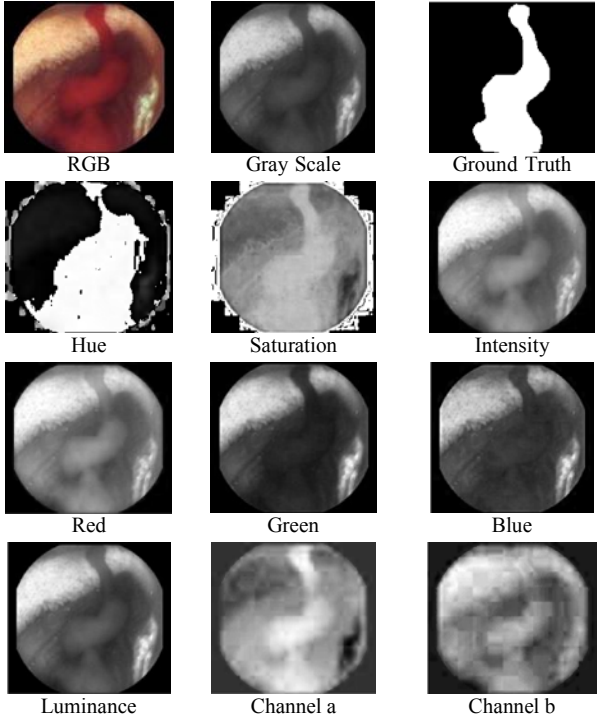
Figure 2. Different color spaces.

phase by using an equal number of bleeding and non-bleeding pixels.

*C. Neural Network quantization*

An artificial neural network structure is consisted of three parts from hardware implementation perspective. Weights are realized as multiplication, input of the activation functions are provided by addition and non-linear activation function can be implemented as a piecewise linear function.

Recently binarization is used as an efficient way to reduce the complexity of the network structure. Simplification of the network structure is performed on all connections. In binarization, a value converts to two possible values for example 0 and 1. Two methods were introduced in previous studies for binarization [20, 21]. Deterministic and stochastic binarization of the network weights are shown in (1) and (2) respectively.

$$W_b = \begin{cases} +1 & if\ W \geq 0, \\ -1 & otherwise. \end{cases} \quad (1)$$

$$W_b = \begin{cases} +1 & with\ probability\ p = \sigma(w), \\ -1 & with\ probability\ 1-p. \end{cases} \quad (2)$$

$$\sigma(x) = clip\left(\frac{x+1}{2}, 0, 1\right) = \max(0, \min(1, \frac{x+1}{2})) \quad (3)$$

where $W$ is network weight before quantization and $W_b$ is binarized one and $\sigma(x)$ in (3), is the '"hard sigmoid function". In this paper deterministic ternary quantization of the weights is considered as (4).

$$W_b = \begin{cases} -1 & if\ W < 0 \\ 0 & if\ W = 0 \\ 1 & if\ W > 0 \end{cases} \quad (4)$$

In [22] roughly power consumption of the operations with different representation was presented. With respect to the [22] in comparison with 32-bit representation, binary representation consumes 32 times smaller memory size and 32 times fewer memory accesses. Also it is noted in [22] that, energy consumption of addition and multiplication operations are decreased significantly with less bit length representation of the weights.

### III. EXPERIMENTAL RESULTS

For evaluation of the proposed method, experimental results are performed using TensorFlow framework. A PC equipped with an Intel(R) Core(TM) i7-4790 CPU 4.00 GHz and 32GB of RAM is used for training and testing the network performance. An MLP structure is trained and tested on the WCE bleeding images which are free and available online in [23]. DICE score is used for classification performance and is as follow.

$$DICE = \frac{2TP}{2TP + FP + FN}$$

where TP and TN are the number correctly identified bleeding and non-bleeding pixels, respectively. FP and FN are the number of the incorrectly identified bleeding and non-bleeding pixels, respectively. Patch size is selected to be 5×5 experimentally and an MLP with 40-20-8 hidden layers is applied. In Fig. 3, the visual results of the full-precision MLP are illustrated. It is observed that three aforementioned color spaces with an MLP structure are able to segment the bleeding regions. Also visual results of the Network with quantized weights are illustrated in Fig. 3. It is important to note that our experiment, binarization using RELU activation function lead to undesirable segmentation results. So sigmoid activation function is utilized and then the network connections are quantized. In Table 1, simulation results are compared with other related works on WCE image bleeding segmentation. The related works [8] and [9] in Table 1, have the same dataset as ours. In this experiment, 50 images are used for train and test, and network performance is validated using 5-fold cross validation method.

TABLE 1. Segmentation performance in term of DICE score

|  | [8] | [9] | MLP, Full Precision | MLP, Quantized |
|---|---|---|---|---|
| Method | SVM | SVM | ANN | Quantized ANN |
| Min | 0.66 | 0.62 | 0.68 | 0.63 |
| Max | 0.98 | 0.97 | 0.98 | 0.97 |
| Average | 0.840 | 0.810 | 0.8553 | 0.8403 |

### IV. CONCLUSION

A new method for automatic detection of bleeding regions in WCE images was presented. The most informative color channels were selected and an artificial neural network was applied for segmentation. Detection system was designed with respect to limitation of the hardware resources in WCE. Hence neural network with quantized weights during training phase was utilized without any preprocessing or post-processing. Simulation results show that ternary quantized neural network can segment bleeding regions with acceptable performance.

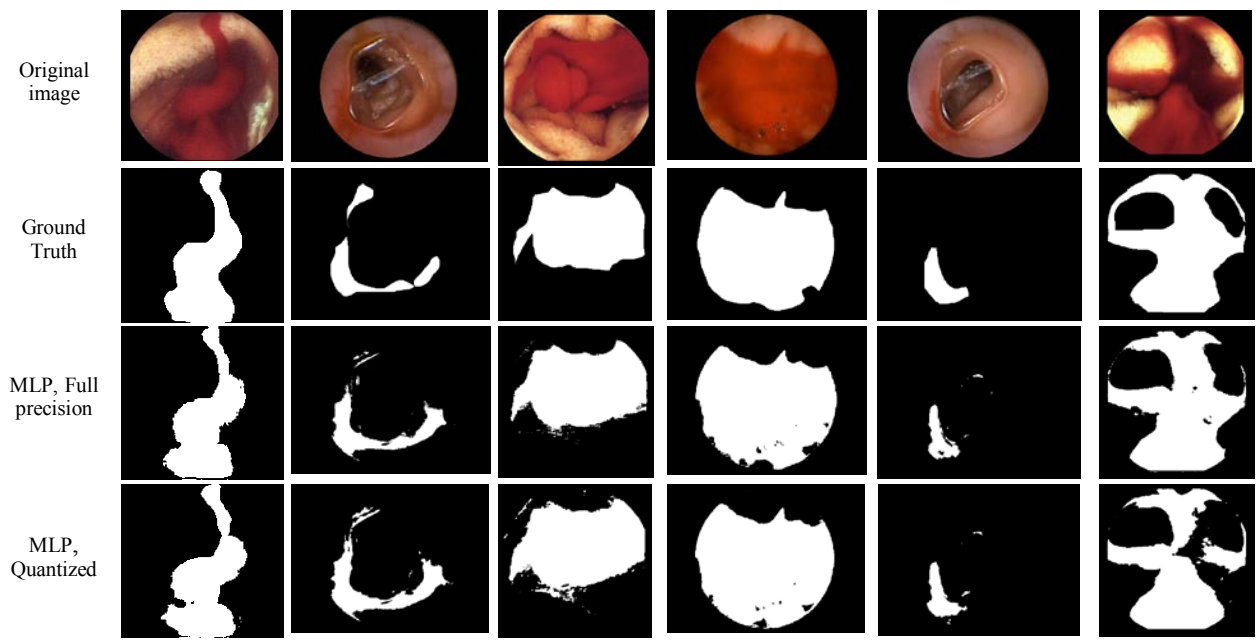

Figure 3. Visual results of bleeding segmentation.

Also no significant drop of accuracy was occurred in comparison with the full-precision network. Quantized neural network without any multiplication could be considered as automatic diagnostic approach inside the capsule.